%% file: bmvc_review.tex
\documentclass{bmvc2k}
\usepackage{url}
\usepackage{amsmath,amssymb,graphicx}
\usepackage{hyperref}
\usepackage{booktabs}
\usepackage{multirow}
\usepackage{mathtools}
\usepackage{cleveref}
\usepackage{colortbl}
\usepackage[export]{adjustbox}
\usepackage{tikz}
\usetikzlibrary{spy}
\usepackage{diagbox}
\usepackage{makecell}
\usepackage{amssymb}
\usepackage{enumitem}
\usepackage{cleveref}


\title{HDRSplat: Gaussian Splatting for High Dynamic Range 3D Scene Reconstruction from Raw Images}

\addauthor{Shreyas Singh${^*}$}{itsshreyas1803@gmail.com}{1}
\addauthor{Aryan Garg${^*}$}{aryangarg019@gmail.com}{1}
\addauthor{Kaushik Mitra}{kmitra@ee.iitm.ac.in}{2}

\addinstitution{Indian Institute of Technology, Madras}

\runninghead{Singh}{HDRSplat}

\def\etal{\emph{et al}\bmvaOneDot}

\begin{document}

\maketitle

\let\thefootnote\relax\footnotetext{* denotes equal contribution}
\input{sections/abstract}
\input{sections/intro}

\input{sections/related}
\input{sections/method}

\input{sections/experiments}

\input{sections/app}
\input{sections/conclusion}

\bibliography{egbib}
\end{document}

%% file: sections/abstract.tex
\begin{abstract}
The recent advent of 3D Gaussian Splatting (3DGS) has revolutionized the 3D scene reconstruction space enabling high-fidelity novel view synthesis in real-time. However, with the exception of RawNeRF, all prior 3DGS and NeRF-based methods rely on 8-bit tone-mapped Low Dynamic Range (LDR) images for scene reconstruction. Such methods struggle to achieve accurate reconstructions in scenes that require a higher dynamic range. Examples include scenes captured in nighttime or poorly lit indoor spaces having a low signal-to-noise ratio, as well as daylight scenes with shadow regions exhibiting extreme contrast. Our proposed method HDRSplat tailors 3DGS to train directly on  14-bit linear raw images in near darkness which preserves the scenes' full dynamic range and content.
Our key contributions are two-fold: Firstly, we propose a linear HDR space-suited loss that effectively extracts scene information from  noisy dark regions and nearly saturated bright regions simultaneously, while also handling view-dependent colors without increasing the degree of spherical harmonics.
Secondly, through careful rasterization tuning, we implicitly overcome the heavy reliance and sensitivity of 3DGS on point cloud initialization. This is critical for accurate reconstruction in regions of low texture, high depth of field, and low illumination. HDRSplat is the fastest method to date that does 14-bit (HDR) 3D scene reconstruction in $\le$15 minutes/scene  ($\sim$30x faster than prior state-of-the-art RawNeRF). It also boasts the fastest inference speed at $\ge$120fps. We further demonstrate the applicability of our HDR scene reconstruction by showcasing various applications like synthetic defocus, dense depth map extraction,  and post-capture control of exposure, tone-mapping and view-point. Source code available at~\href{https://github.com/shreyesss/HDRSplat.git}{https://github.com/shreyesss/HDRSplat}.

\end{abstract}
\begin{figure}[ht]
\includegraphics[width=\textwidth, trim={20pt 20pt 20pt 20pt}, clip]{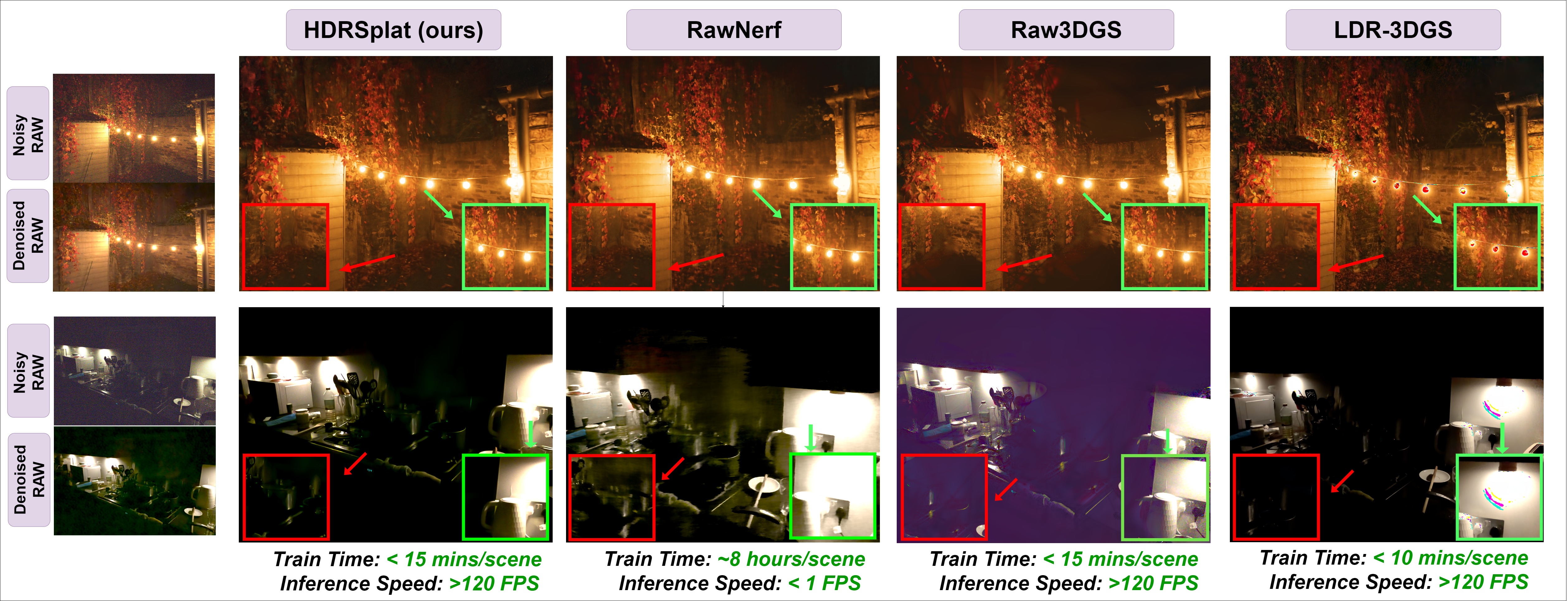} 
    \centering
    \setlength{\abovecaptionskip}{-20pt}
    \caption{\textbf{Qualitative comparison.} 
     \textbf{HDRSplat} enables the highest fidelity 3D scene reconstruction at inference speed of over 120\textit{fps}. This is in contrast with our baselines: \textbf{Raw3DGS} (trained  on demosaiced raw images) \& ~\textbf{LDR-3DGS} (trained on 8-bit LDR images), which give poor quality renders in nighttime scenes and \textbf{RawNeRF}~\cite{rawnerf} which even though similar in fidelity to ours takes 8 hours/scene to train.}
    \label{fig:fig_1_qual_comp}
\end{figure}
~\vspace{-35 pt}

%% file: sections/intro.tex
\section{Introduction}
\label{sec:intro}
Since 2022, Kerbl~\etal's seminal work on real-time high-fidelity scene reconstruction: 3D Gaussian Splatting (3DGS~\cite{3dgs}), has spawned a myriad of impactful  applications, ranging from critical industries like health, urban planning, ~\cite{vastgaussian, HO-gaussian, hugs, urban-gaussian} and robotic-navigation~\cite{navigation_1} to the entertainment and fashion industry~\cite{gen1, gen3, gen5, stylegaussian, edit1, edit2}. 
Despite these advancements, nighttime 3D reconstruction remains a largely unexplored challenge. 
While existing 3DGS~\cite{3dgs} methods excel in well-lit environments, addressing nighttime scenarios is vital for a variety of applications like emergency response mapping, nocturnal security, military operations, urban planning, astronomy, and archaeological documentation. 
However, low textures, photon-shot noise and varying local illumination in the nighttime scenes present formidable challenges, enticing the need for a real-time high-fidelity algorithm in this space.

Traditional 3D reconstruction predominantly relies on 8-bit low dynamic range (LDR) images as input, which does not effectively capture nighttime or low-exposure scenes with varying illumination. 
This limitation manifests in various challenges like low peak signal-to-noise ratio in darkness, heightened contrast, and saturation in regions of extreme brightness. 
Furthermore, the conversion process from raw to 8-bit quantized LDR images introduces loss of valuable information.  
 Midenhall~\etal~\cite{rawnerf} introduced an approach to generate high fidelity novel views directly in the 14-bit linear HDR space, leveraging supervision from noisy raw inputs, demonstrating that High Dynamic Range (HDR) images are needed to faithfully represent such scenes.
However, RawNeRF~\cite{rawnerf}'s high memory consumption accompanied by unrealistic training time of $\sim$8 hours per scene  on a single GPU, makes it impractical for real time use cases.

Our method, HDRSplat, presents a novel adaptation of 3DGS tailored for reconstructing scenes in the same 14-bit linear HDR space. 
Leveraging supervision on denoised raw images, HDRSplat achieves 3D scene reconstruction in under $\sim$15 minutes as shown in~\cref{fig:fig_1_qual_comp}. 
Our approach also overcomes lossy and often irreversible propriety ISP post-processing thus retaining the full dynamic range and scene information. 
Recent literature~\cite{deblur-gaussians, HO-gaussian} has also highlighted 3DGS's inherent limitation in reconstructing high depth of field scenes with low texture and/or low exposure. 
This challenge stems from the over reliance of the3DGS algorithm on the quality of point cloud obtained from COLMAP's SfM~\cite{SFM_COLMAP}. THis often results in non-uniform point density in regions of bad initialization leading to  blurriness, color hallucinations, and other artifacts during reconstruction. 
Through a task-specific loss function design in HDR space, careful rasterization tuning, and  deep Bayer space denoising, we address all these issues, thus achieving  state-of-the-art HDR reconstructions while also significantly improving the point cloud quality (\cref{fig:point-cloud-holes}).
HDRSplat further enables novel view synthesis and post-capture control for these views like focus manipulation, exposure control and custom tonemapping for dynamic range compression.
Additionally, the accurate mesh extraction from the 3DGS point cloud enabled by Gu{\'e}don~\etal~\cite{sugar} enables further downstream creative XR-applications like 3D scene editing, virtual object insertion \& deletion. To summarize, we make the following four key contributions:

\begin{itemize}[itemsep=0pt]
    \item[\textbf{--}] We propose HDRSplat, the first 3D Gaussian Splatting adaptation for dense scene reconstruction in linear HDR space leveraging supervision directly from raw sensor data. 
    
    \item[\textbf{--}] HDRSplat achieves the \textbf{fastest 3D scene reconstruction in linear HDR space} in $\le$15 minutes ($\sim$30X faster than prior state-of-the-art RawNeRF) at an inference speed of $\ge$120 \textit{fps}. 
    
    \item[\textbf{--}] Through our proposed combination of a \textbf{novel loss function}, \textbf{Bayer-space denoising, and rasterization tuning} we achieve superior fidelity ((4\% LPIPS($\downarrow$), 4\% SSIM($\uparrow$), 0.5 dB PSNR($\uparrow$)) over prior state-of-the-art RawNeRF.

    \item[\textbf{--}] We showcase the advantages of rendering in linear HDR space by \textbf{demonstrating applications} such as synthetic defocus, custom tonemapping, exposure adjustment and dense depth map extraction.

\end{itemize}
~\vspace{-25 pt}

%% file: sections/related.tex
\section{Related Work}
\label{sec:related}
\textbf{3D Gaussian Splatting:}
3DGS~\cite{3dgs} has found extensive usage in crucial medical applications like endoscopy~\cite{endoscopy_3dgs}, computed tomograpy~\cite{ct_3dgs}, and X-ray imaging~\cite{xray_3dgs}, to scalable techniques~\cite{vastgaussian, HO-gaussian}, urban navigation~\cite{hugs},~\cite{urban-gaussian} deformable object reconstruction~\cite{deformable_3dgs}, SLAM~\cite{slam_1, slam_2, slam_3}, LiDAR processing~\cite{lidar}, robotic navigation~\cite{navigation_1}, and generative 3D applications~\cite{gen1, gen2, gen3, gen4, gen5} ranging from style transfer~\cite{stylegaussian} to 3d-editing~\cite{edit1, edit2}. 
Even applications including mirror~\cite{mirror} and reflection reconstruction~\cite{reflective}, relighting~\cite{relight}, and mesh extraction~\cite{sugar} for XR have been explored. 
Still, nighttime reconstruction using 3DGS remains largely unexplored due to the inherent challenges and constraints of the problem. 
Our low-light conditions exacerbate issues such as improper point cloud initialization, limited and blurred information, ambiguous scene depth, and overall high dynamic range.

\noindent \textbf{Stabilizing 3DGS:}
\label{subsec:stablize3dgs}
Many methods have explored regularization or stabilization for 3DGS to increase its robustness for the reconstruction of blurred and depth-ambiguous (horizon) scenes. 
RAIN-GS~\cite{rain_gs} uses a sparse-large-variance random initialization. Approaches such as HO-Gaussian~\cite{HO-gaussian} utilize grid-based volumes with MLPs to sample points in under-represented regions, which improves the optimization of geometric information in urban scenes. Similarly, Deblurring Gaussians~\cite{deblur-gaussians} leverage a uniform distribution over the 3D space to sample new points at fixed intervals, it also employs a simple KNN strategy to interpolate the properties of the newly sampled points for densification. 
Trilinear Point Splatting~\cite{trips} explores a feature pyramid-based rendering pipeline. 
However, these 3DGS-based stabilization methods operate under the \textit{strong assumption} of well-lit and low-dynamic range conditions. 
Hence, they remain largely unsuitable for nighttime 3D HDR reconstructions.
\noindent \textbf{Night Time HDR 3D Reconstruction:}
So far, only one work, to the best of our knowledge, has tackled nighttime 3D reconstruction: RawNeRf~\cite{rawnerf}.
However, this is an implicit neural radiance field~\cite{nerf} based method relying on accurate yet extremely time-consuming ray-marching. 
RawNeRF renders in the raw space using a Mip-NeRF~\cite{mipnerf} backbone unlike our denoised-HDR renderings with 3DGS as the backbone. 
This allows both our methods to apply custom tonemapping and develop downstream XR, postprocessing, etc. applications but our method renders at 100+ fps (30x faster training) with similar if not superior fidelity.

\noindent \textbf{Denoising:}
Removal of photon shot noise from raw sensor output is an integral step of our end-to-end pipeline. 
We empirically test a set of traditional filtering-based denoisers: BM3D~\cite{BM3D}, median, and bilateral. 
We also use evaluate a deep neural network based denoiser introduced by Wang~\etal (PMRID~\cite{PMRID}).
It provides the highest fidelity results (\cref{fig:bayer_space_denoise}) and successfully recovers scene information from noisy low illumination regions.

~\vspace{-15 pt}

%% file: sections/method.tex
~\vspace{-20 pt}
\section{HDRSplat}

\begin{figure}[h]
\includegraphics[width=0.99\textwidth, clip]{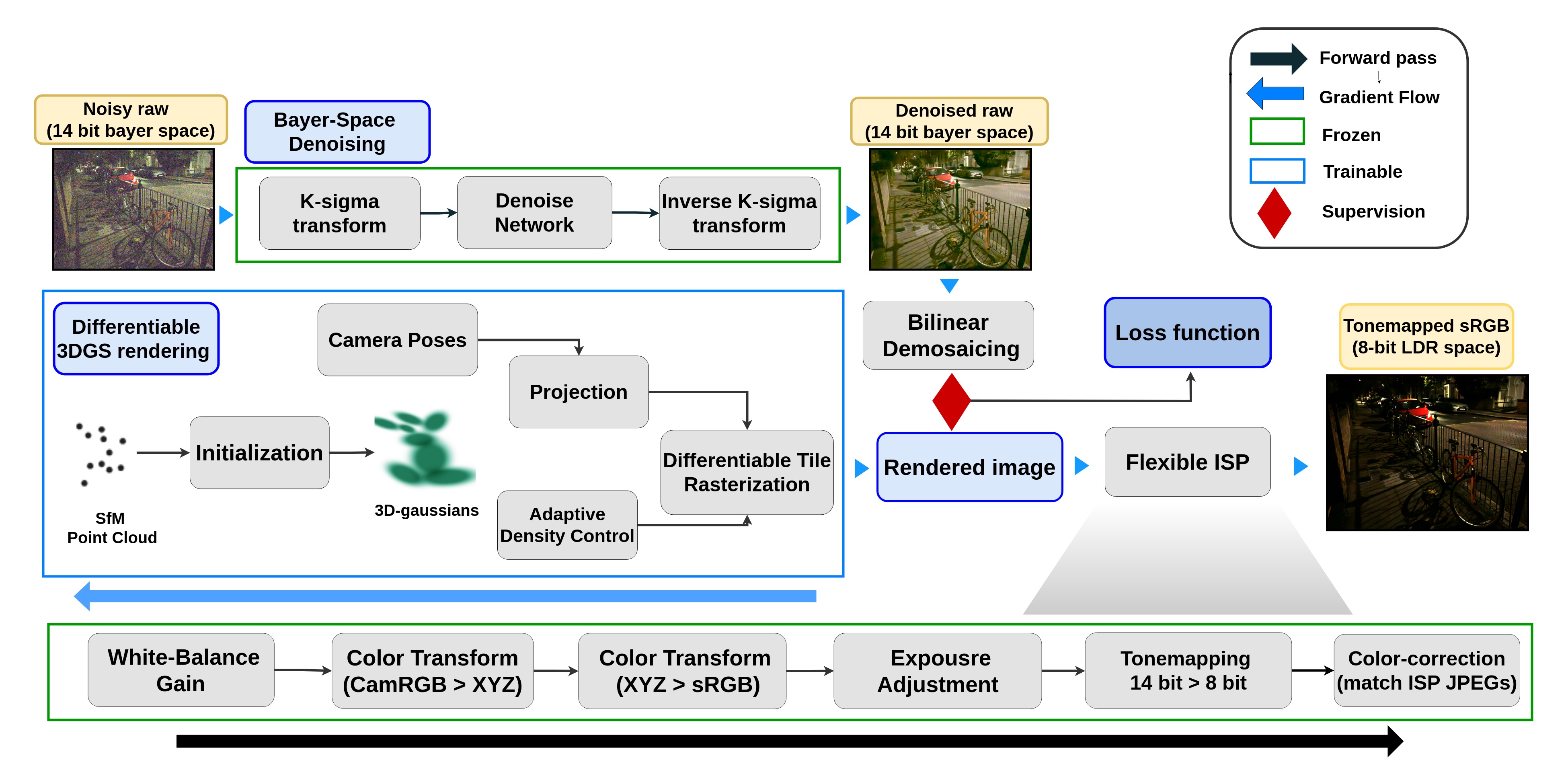} 
    \centering
    \setlength{\abovecaptionskip}{-10pt}
    \caption {\textbf{Rendering Pipeline} for generating novel HDR views from noisy raw inputs. The 3 key components highlighted are: (1) \textbf{Bayer-space-denoising} (2) \textbf{Differentiable 3DGS rasterization} (3) \textbf{Flexible ISP} to convert from 14-bit linear raw to  tonemapped 8-bit sRGB.}
    \label{fig:pipeline}
\end{figure}

Our end-to-end pipeline for 3D scene reconstruction consists of 3 key stages as shown in~\cref{fig:pipeline}. 
First, the pre-processing step involves Bayer-space denoising to remove view-dependent photon shot noise from raw images using PMRID~\cite{PMRID}. 
This is followed by a bilinear demosaicing step, which outputs a 3-channel 14-bit linear raw image with a reduced per-pixel noise level. 
The subsequent stage uses the denoised, demosaiced linear raw images to optimize the parameters of 3D Gaussians with our stop-gradient ($sg(.)$) scaled $\mathcal{L}_1 + DSSIM$ loss. 
We tailor the rasterization parameters during training to implicitly handle the initialization sparsity and non-uniformity of the generated point cloud.
Our final step involves a minimalist, fully flexible post-processing pipeline. This converts the novel views from 14-bit linear raw space to 8-bit tonemapped sRGB space, enabling manipulation of focus, tone-mapping, and exposure for the rendered views.


\label{sec:method}
~\vspace{-15 pt}
\subsection{Background: 3DGS - Differentiable Rendering}
Our Gaussian Splatting ~\cite{3dgs} based renderer uses point clouds to explicitly model a 3D scene. 
Each point in the cloud is a 3D Gaussian.
According to Zwicker~\etal~\cite{world_coords_3dgs}, each 3D Gaussian ($G_{i}$) in world coordinates is centered at a unique position vector $\mu_i$ and is further defined by an anisotropic 3D covariance matrix ($\Sigma$), and spherical harmonics (SH) to represent view-dependent appearance. 
The input for the proposed method consists of camera poses and point clouds, which can be obtained through the structure from motion (SfM), and a collection of images.
~\vspace{-4 pt}
\begin{equation}
    G_i(x) = e^{-\frac{1}{2}(\mathbf{x} - \mu_i)^T \Sigma^{-1}_i (\mathbf{x} - \mu_i)}.
    \label{eq:g_in_3d_world}
\end{equation}
Zwicker~\etal~\cite{world_coords_3dgs} also demonstrate the projection to image space from a pose represented by the viewing transformation $\mathcal{W}$, the projected covariance $\Sigma^{i}_i$ as follows:
\begin{equation}
    \Sigma^{'}_i = \mathcal{J} \mathcal{W} \Sigma_i \mathcal{W}^T \mathcal{J}^T.
    \label{eq:projected_covariance}
\end{equation}
where $\mathcal{J}$ is the Jacobian of the local affine approximation of the projective transformation. Kerbl \etal~\cite{3dgs} reformulated the covariance matrix ($\Sigma$ = $\mathcal{R}\mathcal{S}\mathcal{S}^T\mathcal{R}^T$) using scaling ($\mathcal{S}$) and quaternion matrices ($\mathcal{R}$) to satisfy the positive semi-definiteness constraint during optimization. 
Finally, each Gaussian ($G_i$) also has spherical harmonic (SH) coefficients~\cite{plenoxels} to model view-dependent colors and a learned opacity scalar $\alpha_i$. After Radix-sorting~\cite{radix_sort} the Gaussians per tile, the final rendering equation at each pixel $r$ to have a color $C(r)$ is given by:
~\vspace{-4 pt}
\begin{equation}
    C(r) = \sum_{i=1}^{N} c_i \alpha_i G^{'}_i(r) \prod_{j=1}^{i-1} (1-\alpha_j G^{'}_j(r))
    \label{eq:color_rendering}
\end{equation}
where $N$ is the total gaussians in the scene, $c_i$ is the color of each gaussian and $G^{'}_i(r)$ is the $i^{th}$ 3D gaussian projected to the 2D image space.

\subsection{Optimization}
\noindent \textbf{Bayer Space Denoising:}
Motivated by the need to enhance the supervising signal for the differentiable rasterization module, we integrate PMRID~\cite{PMRID} ($\mathcal{G}_{\theta}$) as a pre-processor to remove photon shot noise in the Bayer raw space.
PMRID's standout feature lies in its use of a k-sigma transform (\cref{fig:pipeline}) to map noisy images captured under different ISO settings into an ISO-invariant signal-noise space. 
This approach enables a compact network trained in this ISO-invariant space to generalize and effectively denoise (varying noise) raw images.
~\vspace{-15 pt}
\begin{figure}[h]
\includegraphics[width=\textwidth, clip]{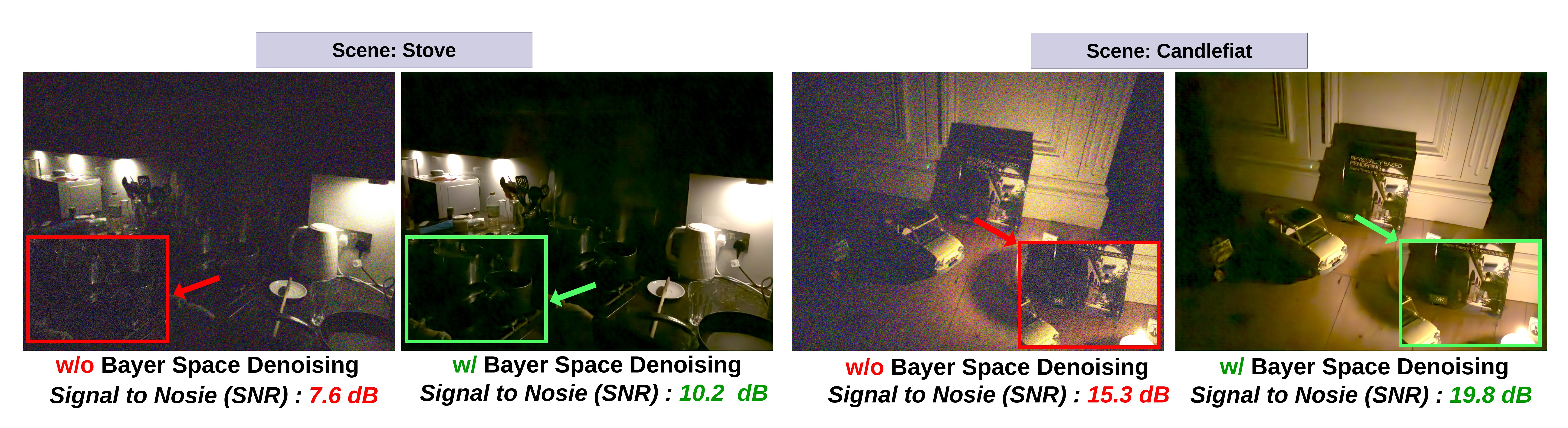}
    \centering
    \setlength{\abovecaptionskip}{-25pt}
    \caption{\textbf{Importance of bayer space denoising using PMRID:} Denoised raw images successfully retrieve scene information from noisy low illumination and saturated bright regions in the raw image space due to a lowered photon shot noise level.}
\label{fig:bayer_space_denoise}
\end{figure}

\textbf{Loss Function:}
Our loss choice is inspired by Noise2Noise's (Lehtinen~\etal~\cite{noise2noise}) stop-gradient ($sg(.)$) scaled $\mathcal{L}_2$ loss for high dynamic range (HDR) reconstruction.
The scaling alleviates long-tail effects present due to unbounded luminance values in HDR samples that lead to non-convergence. 
$\mathcal{L}_2$ loss ensures that the expectation of noisy targets is equal to the true pixel value when operating in the linear HDR space, which does not hold for non-linearly tonemapped LDR space ($E(\mathcal{T}(x)) \neq \mathcal{T}(E(x))$ where $\mathcal{T}$ is a non-linear tonemapping-operator). 
It is statistically trivial to prove that clean targets can be achieved using \textit{only} zero-mean centered noise corrupted ground truths (or raw space) by optimizing over a squared loss over a large number of views.
Multi-Layer perceptrons (MLPs) in NeRFs~\cite{nerf, rawnerf, mipnerf} have high representational power to model and remove per-scene noise implicitly, even with a low number of views.  
This enables Noise2Noise's~\cite{noise2noise} $\mathcal{L}_2$ formulation to work for radiance field methods. 
However, the explicit parameterization of 3D Gaussians lowers the representational power of 3DGS limiting its ability to learn true pixel values from noisy targets that have view-dependent photon-shot noise. 
Hence, we can not use the same $\mathcal{L}_2$ formulation for our method as it results in biased pixel distribution estimation and poorly learned SH coefficients. 
So, we incorporate a pre-trained bayer-space denoising network ($\mathcal{G}_{\theta}$) to eliminate this view-dependent photon-shot noise from the raw input, and a stop-gradient scaled $\mathcal{L}_1$ loss (median expectation) to ensure sharp renders from denoised raw input. 
Additionally, we add a D-SSIM loss term, similar to 3DGS, to improve the structural quality of the renders. Overall, our loss is given by:

%
%
\begin{equation}
    \mathcal{L}_{total} = \lambda \left[ \frac{\| \hat{y} - \mathcal{G}_{\theta} (y_{raw}) \|_1}{(sg(\hat{y}) + \epsilon)} \right]   + (1 - \lambda) DSSIM(\hat{y}, \mathcal{G}_{\theta}(y_{raw})).
    \label{eq:loss_ours}
\end{equation}





\noindent \textbf{Rasrerization tuning:} 
3DGS faces inherent challenges when dealing with scenes characterized by low texture variation, low illumination, and/or high depth of field, both in LDR and HDR settings. 
This is attributed to the limitations of the adaptive densification and pruning module in 3DGS, which only provides localized density control over the point cloud. 
Specifically, splitting and pruning of Gaussians occur primarily in and around regions with dense SfM initialization~\cite{SFM_COLMAP}. 
Consequently, areas with poor SfM initialization suffer from severe under-reconstruction due to the absence of Gaussians that explicitly represent such regions, as shown in 
\cref{fig:point-cloud-holes}:
\begin{figure}[h]
\includegraphics[width=\textwidth]{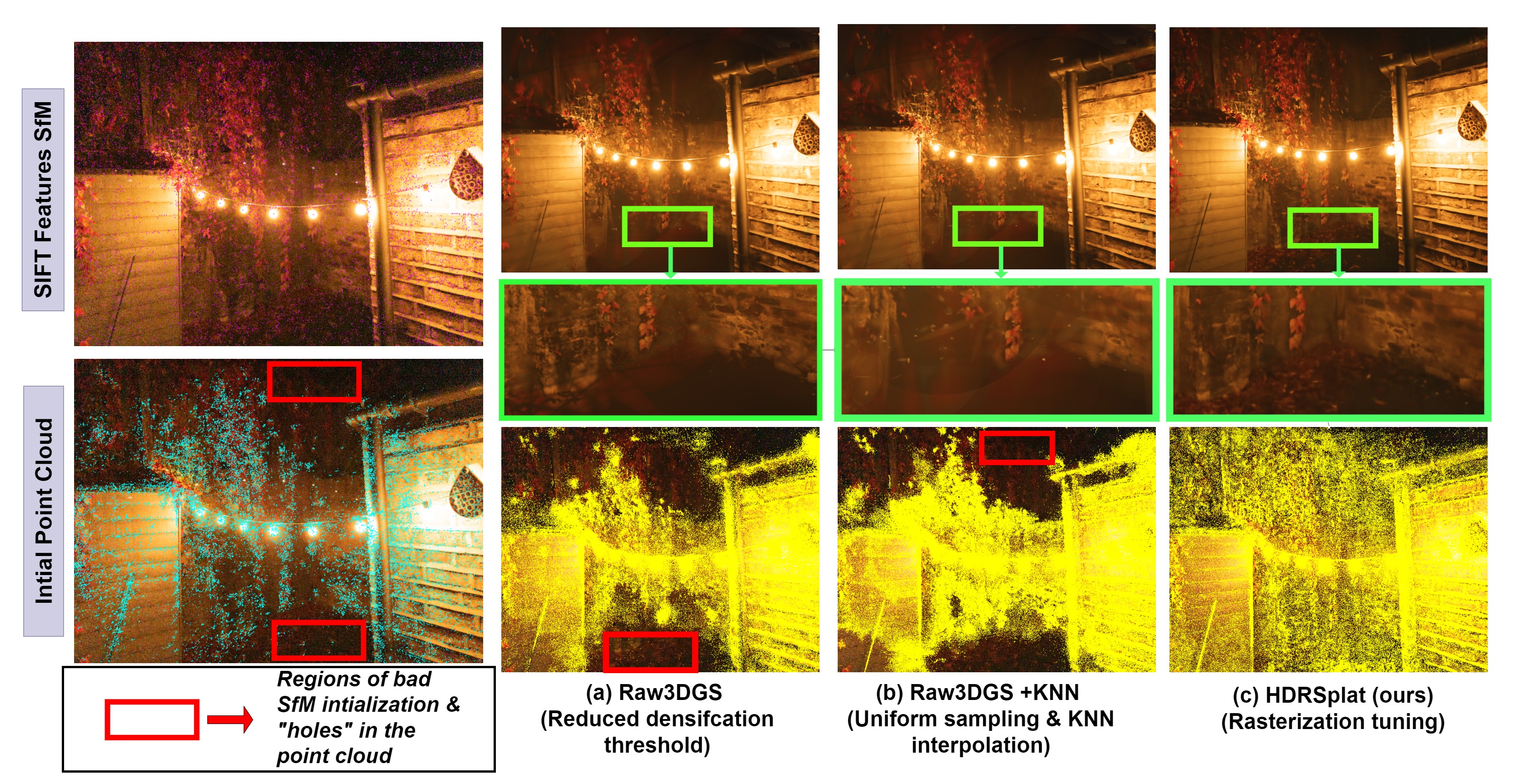}
    \centering
    \setlength{\abovecaptionskip}{-25pt}
    \caption {\textbf{Point Cloud Densification}: (a-b) highlight the non-uniformity and sparsity in regions of poor SfM initialization, leaving \textit{holes} in the final point cloud. (c) shows improved reconstruction via our rasterization method addressing the under-reconstruction problem}
\label{fig:point-cloud-holes}
\end{figure}

\noindent By intuitively adjusting rasterization parameters, we show that avoiding Gaussian pruning in the initial training phase effectively addresses the issue of under-reconstruction caused by the sparse and non-uniform nature of the point cloud. This simple approach contrasts with methods that forcefully augment points in the cloud, as discussed in \cref{subsec:stablize3dgs}.
We achieve this by halving the learning rate for scaling and setting the translation learning rate to 1/5 of the original value for all 3D Gaussians.
This strategic adjustment limits the expansion and movement of Gaussians, preventing them from growing excessively large and overfitting distant sparse or uninitialized regions, or deviating significantly from their original positions, which can lead to pruning in those regions.
Consequently, this \textit{simple regularization} overcomes blurriness, floaters, and under-reconstruction in areas with poor SfM initialization.

%% file: sections/experiments.tex
\vspace{-4mm}
\section{Experiments \& Results}
\label{sec:experiments}
~\vspace{-20 pt}
\begin{table}[h]
    \centering
    \small
    \setlength{\tabcolsep}{2 pt}
    \begin{tabular}{@{\hspace{15 pt}}l|ccc|ccc|ccc@{\hspace{ 15 pt}}}
    \hline
    {\textbf{Method} ($\rightarrow$)}& \multicolumn{3}{c|}{RawNeRF~\cite{rawnerf}} & \multicolumn{3}{c|}{Raw3DGS} & \multicolumn{3}{c}{\textbf{HDRSplat}} \\
    \cline{2-10}
    {\textbf{Scene} ($\downarrow$)}& PSNR & SSIM & LPIPS & PSNR & SSIM & LPIPS & PSNR $\uparrow$ & SSIM $\uparrow$ & LPIPS $\downarrow$ \\
    \hline\hline
    Bikes  & \cellcolor{black!35}\textbf{29.29} &  \cellcolor{black!25}0.76 &  \cellcolor{black!35}\textbf{0.29} & \cellcolor{black!10}27.10 &  \cellcolor{black!25}0.65 & \cellcolor{black!10}0.38 & \cellcolor{black!25}28.49 & \cellcolor{black!35}\textbf{0.80} & \cellcolor{black!35}\textbf{0.29} \\
    Stove  & \cellcolor{black!25}29.42 &  \cellcolor{black!35}0.92 & \cellcolor{black!35}0.11 & \cellcolor{black!10}27.72 &  \cellcolor{black!25}0.82 & \cellcolor{black!10}0.13 & \cellcolor{black!35}\textbf{31.42} & \cellcolor{black!35}\textbf{0.94} & \cellcolor{black!35}\textbf{0.09} \\
    Parkstatue & \cellcolor{black!10}29.00 &  \cellcolor{black!25}0.79 & \cellcolor{black!25}0.23 & \cellcolor{black!25}29.07 &  \cellcolor{black!25}0.73 & \cellcolor{black!10}0.28 & \cellcolor{black!35}\textbf{30.75} & \cellcolor{black!35}\textbf{0.85} & \cellcolor{black!25}0.21 \\ 
    Sharpshadow   & \cellcolor{black!35}\textbf{27.94} &  \cellcolor{black!35}\textbf{0.80} & \cellcolor{black!35}\textbf{0.21} & \cellcolor{black!10}25.75 &  \cellcolor{black!25}0.74 & \cellcolor{black!10}0.29 & \cellcolor{black!25}26.72 & \cellcolor{black!25}0.77 & \cellcolor{black!25}0.26 \\ 
    Candlefiat  &  \cellcolor{black!25}31.19 & \cellcolor{black!35}\textbf{0.86} & \cellcolor{black!25}0.28 & \cellcolor{black!10}27.48 &  \cellcolor{black!25}0.80 & \cellcolor{black!10}0.29 & \cellcolor{black!35}\textbf{31.96} & \cellcolor{black!35}\textbf{0.86} & \cellcolor{black!25}0.24 \\ 
    Notchbush  & \cellcolor{black!25}27.87 &  \cellcolor{black!35}\textbf{0.76} & \cellcolor{black!10}0.40 & \cellcolor{black!10}27.58 &  \cellcolor{black!25}0.72 & \cellcolor{black!10}0.41 & \cellcolor{black!35}\textbf{29.18} & \cellcolor{black!25}0.75 & \cellcolor{black!25}0.36 \\ 
    Nightstreet   & \cellcolor{black!10}28.07 &  \cellcolor{black!25}0.75 & \cellcolor{black!25}0.20 & \cellcolor{black!25}28.12 &  \cellcolor{black!25}0.75 & \cellcolor{black!10}0.25 & \cellcolor{black!35}\textbf{31.50} & \cellcolor{black!35}\textbf{0.87} &\cellcolor{black!25}0.20 \\  
    Morningkitchen  & \cellcolor{black!25}29.55 &  \cellcolor{black!25}0.77 & \cellcolor{black!10}0.28 & \cellcolor{black!10}28.77 &  \cellcolor{black!25}0.77 & \cellcolor{black!10}0.30 & \cellcolor{black!35}\textbf{29.90} & \cellcolor{black!35}\textbf{0.83} & \cellcolor{black!25}0.26 \\  
    Livingroom  & \cellcolor{black!35}\textbf{30.61} &  \cellcolor{black!35}\textbf{0.87} &\cellcolor{black!25} 0.19 & \cellcolor{black!25}29.39 &  \cellcolor{black!25}0.85 & \cellcolor{black!10}0.22 & \cellcolor{black!10}29.12 & \cellcolor{black!25}0.86 & \cellcolor{black!25}0.18 \\  
    Gardenlights  & \cellcolor{black!35}\textbf{24.43} &  \cellcolor{black!35}\textbf{0.53} & \cellcolor{black!10}0.46 & \cellcolor{black!10}23.08 &  \cellcolor{black!10}0.51 & \cellcolor{black!10}0.46 & \cellcolor{black!10}24.35 & \cellcolor{black!35}\textbf{0.53} & \cellcolor{black!35}\textbf{0.42} \\  
    Scooter   & \cellcolor{black!25}35.10 &  \cellcolor{black!35}\textbf{0.88} & \cellcolor{black!25}0.28 & \cellcolor{black!25}33.22 &  \cellcolor{black!25}0.79 & \cellcolor{black!10}0.33 & \cellcolor{black!35}\textbf{35.36} & \cellcolor{black!35}\textbf{0.88} & \cellcolor{black!35}\textbf{0.26} \\  
    Streetcorner  & \cellcolor{black!25}31.79 & \cellcolor{black!35}\textbf{0.84} & \cellcolor{black!25}0.24 & \cellcolor{black!10}27.87 & \cellcolor{black!25}0.73 & \cellcolor{black!10}0.30 & \cellcolor{black!35}\textbf{32.16} & \cellcolor{black!35}\textbf{0.84} & \cellcolor{black!35}\textbf{0.23} \\  
    \hline \hline
    \textbf{Average} & \textbf{29.52} & \textbf{0.79} & \textbf{0.26} & \textbf{27.92}  & \textbf{0.73}  & \textbf{0.30} &  \textbf{30.07} & \textbf{0.82} & \textbf{0.25} \\  
    \hline 
    \end{tabular}
    \caption{\textbf{Quantitative Comparison} of all models rendering in the 14-bit linear HDR space. We demonstrate superior fidelity of our model over RawNeRF (4\% LPIPS($\downarrow$), 4\% SSIM($\uparrow$), 0.5 PSNR($\uparrow$)) and our baseline Raw3DGS (17\% LPIPS($\downarrow$), 12\% SSIM($\uparrow$), 2.1 dB PSNR($\uparrow$)).}
    \label{tab:quant_comp}
\end{table}
~\vspace{-15 pt}
\subsection{Implementation Details}
\textbf{Dataset:} We use RawNeRF's~\cite{rawnerf} nighttime, 4032$\times$3024 resolution, paired Bayer-raw and proprietary ISP LDR dataset, captured using an iPhone-X. 
The dataset consists of 14 scenes however we omit two scenes: Windowlegovary and Candle due to their multi-exposure captures.
We undistorted the entire dataset into a \textit{simple pinhole} camera model using COLMAP \cite{SFM_COLMAP} since 3DGS does not provide native support for radial camera distortions.

\noindent \textbf{Baselines:} 
We compare HDRSplat with prior state-of-the-art RawNeRF~\cite{rawnerf} and our own constructed baselines: LDR-3DGS and Raw3DGS. 
LDR-3DGS is Kerbl~\etal's original implementation of 3DGS, trained on 8-bit iPhone JPEGs. 
Raw3DGS trains the original implementation on noisy demosaiced raw images with RawNeRF's loss (stop-gradient scaled $\mathcal{L}_2$), highlighting the limitations when directly applying RawNeRF's framework to 3DGS. 
Taking inspiration from~\cite{deblur-gaussians}, Raw3DGS+KNN utilizes a uniform point sampling strategy with KNN-based properties' interpolation for point cloud densification as shown in~\cref{fig:point-cloud-holes}.

\noindent \textbf{Metrics:}
For quantitative benchmarking purposes the rendered novel views in 14-bit demosaiced space are converted to 8-bit tonemapped LDR space using our flexible post-processing pipeline as shown in~\cref{fig:pipeline} and color-corrected using the ground truth LDRs, similar to RawNeRF~\cite{rawnerf}. The metrics reported are SSIM, PSNR and LPIPS (VGG-16)~\cite{lpips}. Refer to the supplementary for exact details on the post-processing pipeline.

\noindent \textbf{Experimental Setup:}
RawNeRF~\cite{rawnerf} follows the original setup. 
In Raw3DGS, we set \textit{densify\_grad\_thresh} to $10^{-4}$ (half of original 3DGS) for maximum densification adaptability to noisy raw inputs. 
HDRSplat trains 3DGS~\cite{3dgs} on PMRID~\cite{PMRID} denoised raw inputs with our stop-gradient ($sg(.)$) scaled $\mathcal{L}_1 + DSSIM$  loss.
We set \textit{scaling\_lr} to $10^{-3}$ (1/5 of the orignal implementation) and \textit{position\_lr\_init} to $8 \times 10^{-5}$ (1/2 of the orignal implementation).
 PMRID~\cite{PMRID} denoising  and bilinear demosaicing steps are pre-computed once per scene and stored, cumulatively  they takes $\sim$ 8 seconds/view to process. 
An exponential decay scheduling similar to~\cite{plenoxels, 3dgs} is used for every 3DGS-based method. 
We use a single NVIDIA RTX 3090 with 24 GB VRAM.

\vspace{-3mm}
\subsection{Results}
The quantitative comparison in~\cref{tab:quant_comp} and compute load~\cref{tab:Fidelity} highlights HDRSplat's capability to generate superior-fidelity renders extremely fast, crucial for real-time applications. 
Notably, HDRSplat achieves better perceptual similarity (LPIPS~\cite{lpips}) compared to LDR-3DGS, which is trained directly on the 8-bit iPhone JPEG ground truths. 
HDRSplat stands out not only for its rapid training and rendering speeds but also for its exceptional memory efficiency (\cref{tab:Fidelity}). 
On average, it utilizes only 0.35M points per scene, compared to 1.5M and 5M points in Raw3DGS and Raw3DGS+KNN, respectively.

\begin{table}[h]
    \centering
    \small
    \setlength{\tabcolsep}{5 pt}
    \begin{tabular}{@{\hspace{15 pt}}l|c|c|c|c|c|c@{\hspace{8 pt}}}
    \hline
    \centering\textbf{Method} & PreP\textsuperscript{*} $\downarrow$ & Training $\downarrow$ & Total $\downarrow$  & Inference & LPIPS $\downarrow$ & \# Points $\downarrow$ \\   
    \hline\hline
    RawNeRF~\cite{rawnerf} & \cellcolor{black!15}\textbf{3 mins} & 8 hrs & 8 hrs & 0.1 FPS & 0.30 & N/A \\ 
    Raw3DGS & \cellcolor{black!15}\textbf{3 mins} & 10 mins & 13 mins &\cellcolor{black!15}\textbf{120 FPS} & 0.30 & 1.5 M \\
    Raw3DGS+KNN & \cellcolor{black!15}\textbf{3 mins} & 20 mins & 23 mins &\cellcolor{black!15}\textbf{120 FPS} & 0.28 & 5 M \\ 
    LDR-3DGS~\cite{3dgs} & N/A & \cellcolor{black!15}\textbf{7 mins} & \cellcolor{black!15}\textbf{7 mins} & \cellcolor{black!15}\textbf{120 FPS} & 0.26 & \cellcolor{black!15}\textbf{ 0.35 M} \\ 
    \textbf{HDRSplat (ours)} & 7 mins &\cellcolor{black!15}\textbf{7 mins} & 14 mins & \cellcolor{black!15}\textbf{120 FPS} & \cellcolor{black!15}\textbf{0.25} & \cellcolor{black!15}\textbf{ 0.35 M} \\  
    \hline
    \end{tabular}
    \caption{\textbf{Real time rendering}: We are 30x faster than RawNeRF in training and can infer at 120 fps. PreP\textsuperscript{*} (Pre-processing) refers to  Bayer-space denoising step and bilinear-demosaicing. All metrics have been calculated for an average of 50 views per scene.}
    \label{tab:Fidelity}

\end{table}

~\vspace{-10 pt}

\begin{figure}[ht]
    \centering
    \includegraphics[width=\textwidth]{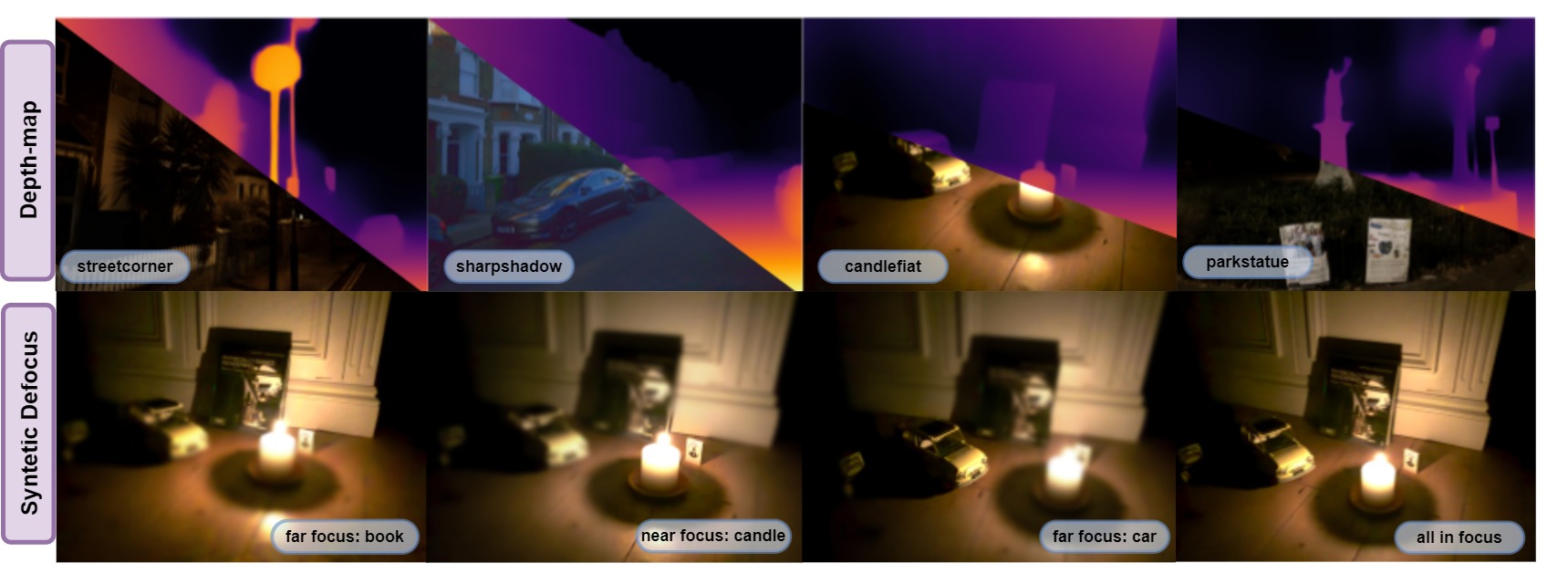}
    \setlength{\abovecaptionskip}{-25pt}
    \caption{\textbf{Application: }Accurate Depth from Novel Views using~\cite{depthanything} and Synthetic-Defocus}
    \label{fig:apps1}
\end{figure}

~\vspace{-20 pt}
\subsection{Ablations}
In~\cref{tab:ablation} we validate the impact of each of our 3 innovations, rasterization tuning, Denoising ($\mathcal{G}_{\theta}$) and our loss on the rendering quality, speed \& efficiency of HDRSplat.
Directly training 3DGS on demosaiced raw images (Raw3DGS) consumes $\ge 4 \times$ more memory and yields noticeably inferior renders (\cref{fig:fig_1_qual_comp}). 
Our rasterization tuning method substantially reduces the memory usage by $\sim$60\%. Bayer denoising and our proposed loss function seem to have the highest impact on perceptual quality based on LPIPS and SSIM scores.

\begin{table}
\centering
\small
\setlength{\tabcolsep}{3 pt}
\begin{tabular}{@{\hspace{15 pt}}l|c|c|c|c|c|c|c@{\hspace{15 pt}}}
    \hline
    \textbf{Ablation} & Loss & $G_{\theta}$ & $\mathcal{LR}$ & PSNR $\uparrow$  & SSIM $\uparrow$ & LPIPS $\downarrow$ & \# Points $\downarrow$ \\   
    \hline\hline
    Raw3DGS &  $\mathcal{R}$ & $\times$ & $\times$ & 27.92 & 0.73 & 0.30 & 1.5 M \\ 
    w/o Rasterization Tuning& $\mathcal{H}$ & \checkmark & $\times$ & 29.44 & 0.80 & 0.30 & 0.85 M \\
    w/o Bayer Denoising&  $\mathcal{H}$ & $\times$ & \checkmark & 29.49 & 0.78 & 0.27 & \cellcolor{black!15}\textbf{0.25 M} \\ 
    w/o Our Loss Function& $\mathcal{R}$ & \checkmark & \checkmark & 29.97 & 0.80  & 0.26 & 0.30 M  \\

    \textbf{HDRSplat (Ours)} & $\mathcal{H}$ & \checkmark & \checkmark & \cellcolor{black!15}\textbf{30.09} & \cellcolor{black!15}\textbf{0.82} & \cellcolor{black!15}\textbf{0.25} & 0.35 M \\
    \hline
\end{tabular}
\caption{\textbf{Ablation Study} to demonstrate the utility of each component of our pipeline. $\mathcal{H}$ refers to our stop-gradient scaled $\mathcal{L}_1 + DSSIM$ loss, $\mathcal{R}$ refers to RawNeRF's~\cite{rawnerf} stop-gradient scaled $\mathcal{L}_2$. $G_{\theta}$ is  Bayer-space denoising and $\mathcal{LR}$ refers to our rasterization tuning.}
\label{tab:ablation}
\end{table}
\begin{figure}[ht]
    \centering
    \includegraphics[width=0.99\textwidth]{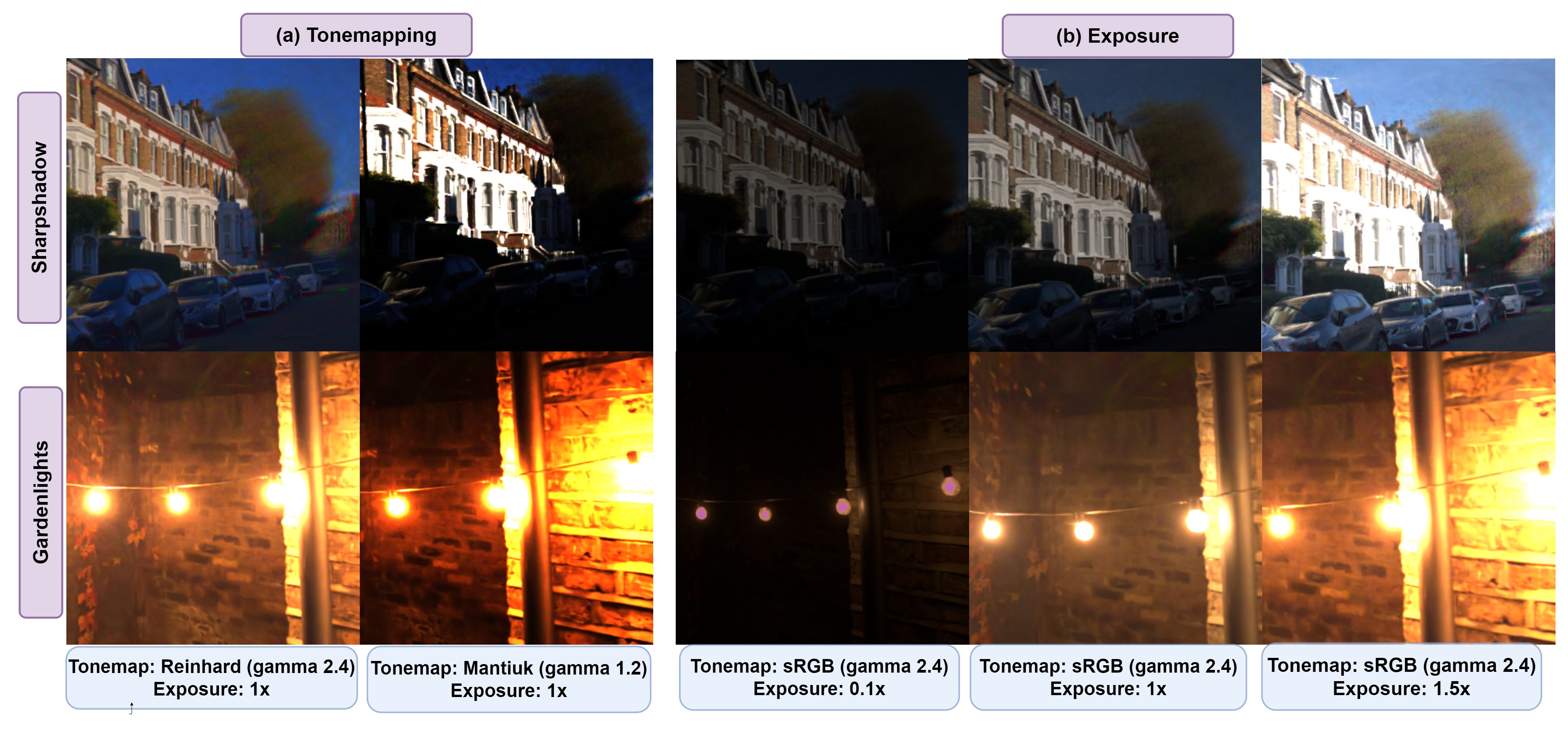}
    \setlength{\abovecaptionskip}{-6pt}
    \caption{\textbf{Application: }HDRSplat, rendering in the linear HDR space, facilitates complete post-capture control for generating visually appealing results in downstream applications.}
    \label{fig:apps2}
\end{figure}
\vspace{-2.2mm}
\subsection{Applications}
In \cref{fig:apps1}, we demonstrate applications such as novel-view synthesis, dense depth map extraction using~\cite{depthanything} and synthetic defocus to highlight regions of interest. All enabled by our method's high-fidelity renders. 
\cref{fig:apps2} illustrates the versatility of variable tonemappings and exposure control made possible by our linear HDR rendering space. These renderings can be retouched like any raw photograph, enabling full post capture control for the user.
\vspace{-2.2mm}






%% file: sections/app.tex


%% file: sections/conclusion.tex
~\vspace{-20 pt}
\section{Conclusion}
\label{sec:conclusion}
\vspace{-2mm}
Our method enables the fastest and highest fidelity HDR scene reconstruction directly from Bayer raw images, while our approach offers significant advantages, such as capturing fine details in high dynamic range scenes and avoiding image compression artifacts, it does present some challenges. 
Dependency on LDR images for SfM~\cite{SFM_COLMAP} and 3DGS' high sensitivity to point cloud initialization, especially in low-light scenarios, could be alleviated in future work by using a COLMAP-free or depth projection based initialization as shown by~\cite{colmap_free}. 
Additionally, designing a noise model or parameterization for individual 3D Gaussians could eliminate the need for external denoising dependencies, streamlining the pipeline and improving overall compute efficiency.
Despite its limitations, HDRSplat represents a significant advancement towards real-time, adaptable, and high-fidelity 3D scene reconstruction, particularly for higher dynamic range scenes.